\documentclass[twocolumn]{article}
\usepackage{amsmath}
\usepackage{graphicx}
\usepackage{hyperref}
\usepackage{authblk}
\usepackage{amssymb} 
\usepackage{titlesec} 

\usepackage{booktabs}
\usepackage{subfig} 
\usepackage[margin=1in, top=1in]{geometry} 

\title{Chebyshev Polynomial-Based Kolmogorov-Arnold Networks: \\
An Efficient Architecture for Nonlinear Function Approximation}

\author[1]{Sidharth SS}
\author[1]{Gokul R}
\author[1]{Anas K P}
\author[2]{Keerthana AR}
\affil[1]{Indian Institute of Science Education and Research Mohali, India}
\affil[2]{Indian Institute of Science Education and Research Thiruvananthapuram, India}

\date{May 2024}

\begin{document}

\twocolumn[
  \begin{@twocolumnfalse}
    \maketitle
    \begin{abstract}
Accurate approximation of complex nonlinear functions is a fundamental challenge across many scientific and engineering domains. Traditional neural network architectures, such as Multi-Layer Perceptrons (MLPs), often struggle to efficiently capture intricate patterns and irregularities present in high-dimensional functions. This paper presents the Chebyshev Kolmogorov-Arnold Network (Chebyshev KAN), a new neural network architecture inspired by the Kolmogorov-Arnold representation theorem, incorporating the powerful approximation capabilities of Chebyshev polynomials. By utilizing learnable functions parametrized by Chebyshev polynomials on the network's edges, Chebyshev KANs enhance flexibility, efficiency, and interpretability in function approximation tasks. We demonstrate the efficacy of Chebyshev KANs through experiments on digit classification, synthetic function approximation, and fractal function generation, highlighting their superiority over traditional MLPs in terms of parameter efficiency and interpretability. Our comprehensive evaluation, including ablation studies, confirms the potential of Chebyshev KANs to address longstanding challenges in nonlinear function approximation, paving the way for further advancements in various scientific and engineering applications.

    \end{abstract}
    \vspace{1cm}
  \end{@twocolumnfalse}
]

\section{Introduction}

In the realm of neural networks, Multi-Layer Perceptrons (MLPs) have long been foundational for approximating nonlinear functions, a capability underpinned by the universal approximation theorem which asserts that MLPs can \cite{1,16}approximate any continuous function given sufficient parameters. Despite their effectiveness, MLPs face notable limitations in terms of parameter efficiency and interpretability. They often require a large number of parameters to achieve high accuracy, and their fixed activation functions make them less adaptable to complex data patterns without extensive post-analysis. Inspired by the Kolmogorov-Arnold representation theorem, Kolmogorov-Arnold Networks (KANs) offer a novel approach by placing learnable activation functions on the edges instead of the nodes\cite{2}, transforming each weight parameter into a learnable univariate function, often parametrized as a spline\cite{5}. This innovative design allows KANs to sum incoming signals at the nodes and apply nonlinear transformations via these learnable functions, enhancing the network's flexibility and adaptability.

Building upon the principles of KANs, the Chebyshev Polynomial-based Kolmogorov-Arnold Network (Chebyshev KAN) incorporates Chebyshev polynomials to parametrize the learnable functions. Renowned for their approximation properties, particularly rapid convergence and numerical stability\cite{3}, Chebyshev polynomials enable Chebyshev KANs to improve the accuracy and efficiency of function approximation tasks. Unlike traditional MLPs where each node applies a fixed nonlinear activation function to the weighted sum of its inputs, Chebyshev KANs employ learnable activation functions on the edges, allowing the network to dynamically adjust its response to input data and model intricate relationships more effectively. The architecture of Chebyshev KANs can be understood as a fully connected network where edges possess learnable functions parametrized by Chebyshev polynomials, defined by their recurrence relation and orthogonality properties. This parametrization leads to high approximation accuracy with fewer parameters, offering a compact representation of complex functions.
 The research demonstrates that Chebyshev KANs surpass original KAN implementations in efficiency\cite{19}, marking a significant advancement in neural network architecture. These networks are particularly suited for approximating complex, high-dimensional functions and have shown superior performance in applications such as solving partial differential equations, making them valuable tools in physics, engineering, and data science.

\vspace{5pt}

\begin{figure}[htbp]
  \centering
  \includegraphics[width=0.5\textwidth,height=6cm]{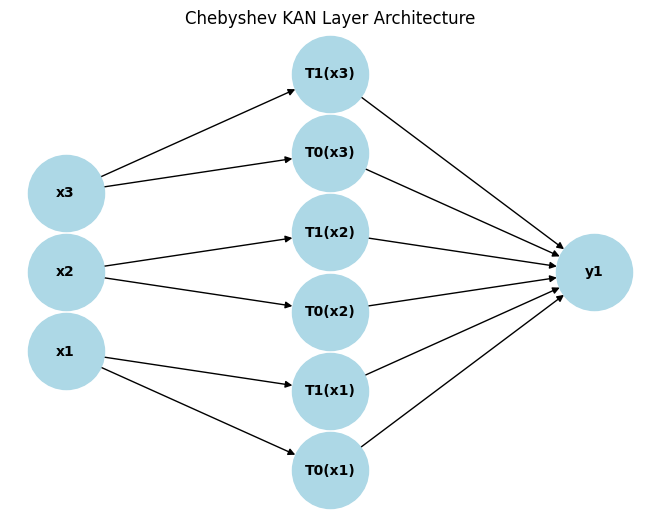}
  \caption{Visualization of the Chebyshev KAN model with 3 input features, degree 1, and output shape 1. The weights/coefficients are not shown in the picture.}
  \label{fig:example}
\end{figure}
 This paper presents a comprehensive investigation of the Chebyshev Kolmogorov-Arnold Network, a transformative approach that seamlessly integrates the theoretical foundations of approximation theory with the computational efficiency of Chebyshev polynomials. Through rigorous theoretical formulations, systematic ablation studies, and extensive experiments on complex functions, we demonstrate the potential of this architecture to address longstanding challenges in nonlinear function approximation, paving the way for further advancements and applications across diverse scientific and engineering domains.

All the experiments and the implementation can be found in this GitHub repository: \href{https://github.com/sidhu2690/ChebyshevKAN}{https://github.com/sidhu2690/ChebyshevKAN}.

\section{Kolmogorov-Arnold Theorem}
The Kolmogorov-Arnold Theorem, also known as the Superposition Theorem or the Kolmogorov-Arnold Representation Theorem, is a fundamental result in approximation theory\cite{4,6}. It states that any continuous multivariate function on a bounded domain can be represented as a superposition (composition) of a limited number of one-variable (univariate) functions and a set of linear operations.
Formally, for a continuous function \( f : [0, 1]^{n} \to \mathbb{R} \) on the \( n \)-dimensional unit hypercube \( [0, 1]^{n} \), the Kolmogorov-Arnold Theorem guarantees the existence of continuous univariate functions \( \Phi_q \) and \( \phi_{q,p} \) such that:
\[
f(x_1, x_2, \ldots, x_n) = \sum_{q=0}^{2n} \Phi_q \left( \sum_{p=1}^{n} \phi_{q,p}(x_p) \right)
\]
where \( x = (x_1, x_2, \ldots, x_n) \). Here, \( \phi_{q,p} : [0, 1] \to \mathbb{R} \) and \( \Phi_q : \mathbb{R} \to \mathbb{R} \) are continuous functions.
\vspace{10pt}

The theorem provides a method to break down a complex multivariate function into simpler univariate functions combined through addition. This representation shows that the complexity of a multivariate function can be managed by understanding its behavior in terms of single-variable functions.
\vspace{10pt}

The essence of the theorem lies in its ability to transform a function of many variables into a sum of functions, each depending on only one variable. This decomposition is particularly useful in various fields of mathematics and applied sciences, as it simplifies the analysis and computation of multivariate functions.

The Kolmogorov-Arnold Theorem has profound implications in understanding the structure of continuous functions. It demonstrates that multivariate continuous functions are not fundamentally more complex than univariate ones, as they can be expressed through compositions of the latter.

\section{Chebyshev Polynomials}

Chebyshev polynomials are a sequence of orthogonal polynomials\cite{9,10} that have a wide range of applications in numerical analysis, approximation theory, and solving differential equations. Named after the Russian mathematician Pafnuty Chebyshev, these polynomials are especially important in the context of polynomial approximation\cite{7,8} of functions due to their minimization of the maximum error (minimax property).

\section*{Definition and Equations}

The Chebyshev polynomials of the first kind \( T_n(x) \) and the second kind \( U_n(x) \) are defined in terms of trigonometric functions. 

\subsection*{Chebyshev Polynomials of the First Kind}

The Chebyshev polynomials of the first kind \( T_n(x) \) are defined as:
\[
T_n(x) = \cos(n \arccos(x))
\]

They can also be expressed using the explicit polynomial form:
\[
T_0(x) = 1
\]
\[
T_1(x) = x
\]
\[
T_2(x) = 2x^2 - 1
\]
\[
T_3(x) = 4x^3 - 3x
\]
\[
T_n(x) = 2xT_{n-1}(x) - T_{n-2}(x) \quad \text{for} \quad n \geq 2
\]

\begin{figure}[htbp]
  \centering
  \includegraphics[width=0.5\textwidth,height=6cm]{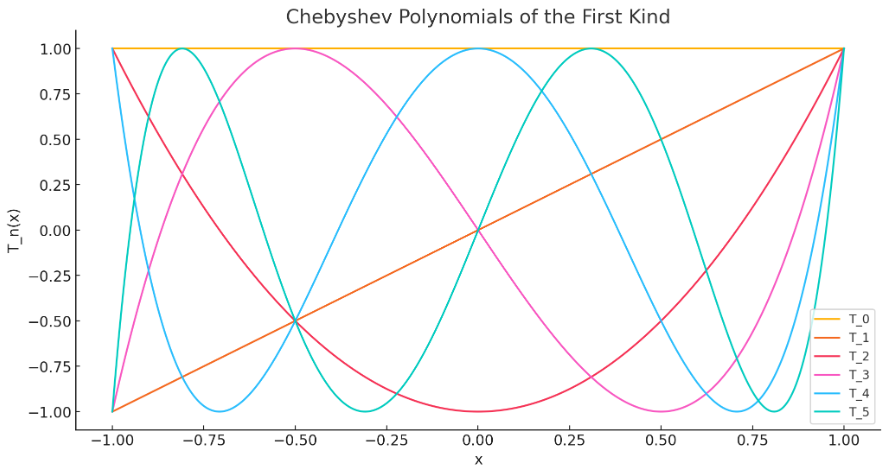}
  \caption{Visualization of the Chebyshev polynomials of the first kind}
  \label{fig:example}
\end{figure}

\subsection*{Chebyshev Polynomials of the Second Kind}

The Chebyshev polynomials of \cite{11}the second kind \( U_n(x) \) are defined as:
\[
U_n(x) = \frac{\sin((n+1) \arccos(x))}{\sqrt{1-x^2}}
\]

Their explicit polynomial forms are:
\[
U_0(x) = 1
\]
\[
U_1(x) = 2x
\]
\[
U_2(x) = 4x^2 - 1
\]
\[
U_3(x) = 8x^3 - 4x
\]
\[
U_n(x) = 2xU_{n-1}(x) - U_{n-2}(x) \quad \text{for} \quad n \geq 2
\]

\begin{figure}[htbp]
  \centering
  \includegraphics[width=0.5\textwidth,height=6cm]{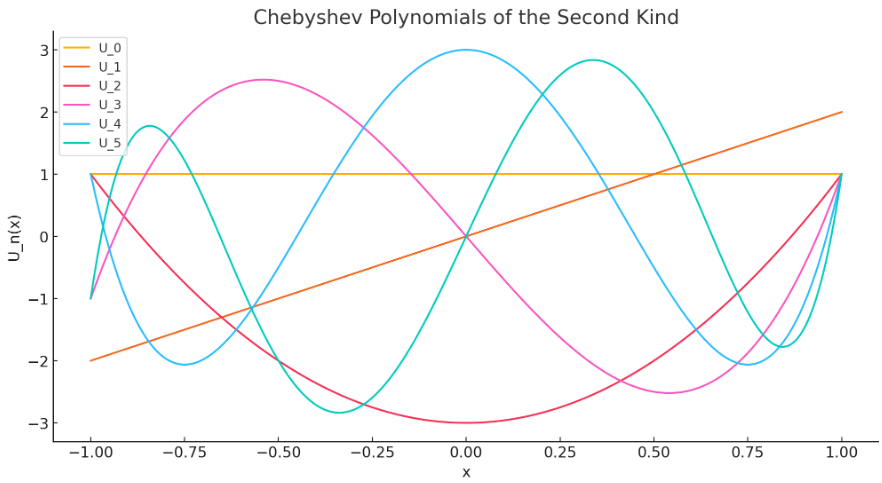}
  \caption{Visualization of the Chebyshev polynomials of the second kind}
  \label{fig:example}
\end{figure}
\section*{Recurrence Relations}

The Chebyshev polynomials satisfy the following recurrence relations, which are useful for their computation:

\subsection*{First Kind}

\[
T_{n+1}(x) = 2xT_n(x) - T_{n-1}(x)
\]

\subsection*{Second Kind}

\[
U_{n+1}(x) = 2xU_n(x) - U_{n-1}(x)
\]

\section*{Orthogonality}

Chebyshev polynomials are orthogonal with respect\cite{3,7} to specific weight functions over the interval \([-1, 1]\):

\subsection*{First Kind}

The polynomials \( T_n(x) \) are orthogonal with respect to the weight function \( \frac{1}{\sqrt{1-x^2}} \):
\[
\int_{-1}^{1} \frac{T_m(x) T_n(x)}{\sqrt{1-x^2}} \, dx = 
\begin{cases} 
0 & \text{if } m \neq n \\
\frac{\pi}{2} & \text{if } m = n \neq 0 \\
\pi & \text{if } m = n = 0
\end{cases}
\]

\subsection*{Second Kind}

The polynomials \( U_n(x) \) are orthogonal with respect to the weight function \( \sqrt{1-x^2} \):
\[
\int_{-1}^{1} U_m(x) U_n(x) \sqrt{1-x^2} \, dx = 
\begin{cases} 
0 & \text{if } m \neq n \\
\frac{\pi}{2} & \text{if } m = n
\end{cases}
\]

\section*{Properties and Applications}

\subsection*{Extremal Properties}

Chebyshev polynomials of the first kind \( T_n(x) \) have the property that they minimize the maximum deviation from zero among all polynomials of the same degree with leading coefficient 1, which makes them useful in polynomial approximation (Chebyshev approximation).

\subsection*{Roots and Extrema}

The roots of \( T_n(x) \) are given by:
\[
x_k = \cos\left( \frac{2k+1}{2n} \pi \right), \quad k = 0, 1, \ldots, n-1
\]

The extrema of \( T_n(x) \) (i.e., the points where \( T_n(x) = \pm1 \)) are given by:
\[
x_k = \cos\left( \frac{k}{n} \pi \right), \quad k = 0, 1, \ldots, n
\]

\section{The Chebyshev Kolmogorov-Arnold Network}

The Chebyshev Kolmogorov-Arnold Network (Chebyshev KAN) is an innovative architecture designed to improve the efficiency and accuracy of nonlinear function approximation. Inspired by the Kolmogorov-Arnold theorem and the powerful approximation capabilities of Chebyshev polynomials, Chebyshev KAN offers significant advantages over traditional Multi-Layer Perceptrons (MLPs).

\subsection{Chebyshev Polynomial Representation}

At the core of Chebyshev KAN is the representation of input data through Chebyshev polynomials. These polynomials are a sequence of orthogonal functions\cite{10} that serve as a robust basis for function approximation.

Given an input tensor $\mathbf{x} \in \mathbb{R}^{\text{batch\_size} \times \text{input\_dim}}$, the network computes a tensor of Chebyshev polynomials $\mathbf{T} \in \mathbb{R}^{\text{batch\_size} \times \text{input\_dim} \times (\text{degree} + 1)}$, where $\text{degree}$ denotes the degree of the Chebyshev polynomial.

The Chebyshev polynomials $T_n(x)$ are defined recursively as:
\[
T_0(x) = 1, \quad T_1(x) = x
\]
\[
\quad T_n(x) = 2x T_{n-1}(x) - T_{n-2}(x) \quad \text{for} \quad n \geq 2.
\]

These polynomials are computed for each input feature across the batch, resulting in a tensor $\mathbf{T}$ that encodes the input data in terms of Chebyshev polynomial bases.

\subsection{Learnable Chebyshev Coefficients}

To enable the network to learn from data, we introduce a tensor of learnable Chebyshev coefficients $\mathbf{C} \in \mathbb{R}^{\text{input\_dim} \times \text{output\_dim} \times (\text{degree} + 1)}$. These coefficients act as parameters that the network adjusts during training to optimize the function approximation.

\subsection{Network Computation}

The network output is computed by performing an Einstein summation operation (einsum) over the Chebyshev polynomials tensor $\mathbf{T}$ and the Chebyshev coefficients tensor $\mathbf{C}$. This operation effectively combines the polynomial bases with the learned coefficients to produce the final output tensor $\mathbf{y} \in \mathbb{R}^{\text{batch\_size} \times \text{output\_dim}}$.

Mathematically, the output tensor $\mathbf{y}$ is computed as:
\[
\mathbf{y}_{bo} = \sum_{i=1}^{\text{input\_dim}} \sum_{j=0}^{\text{degree}} \mathbf{T}_{bij} \cdot \mathbf{C}_{ioj},
\]
where $b$ indexes the batch, $i$ indexes the input dimensions, $o$ indexes the output dimensions, and $j$ indexes the degree of the Chebyshev polynomials.

\subsection{Mathematical Explanation}

The Chebyshev polynomials tensor $\mathbf{T}$ and the Chebyshev coefficients tensor $\mathbf{C}$ are designed to capture the complex relationships between the input and output variables. The einsum operation ensures that each element of the output tensor $\mathbf{y}$ is a weighted sum of the polynomial bases, with weights provided by the learnable coefficients.

\textbf{Example:} Given a batch size of 3, input dimension of 2, output dimension of 3, and polynomial degree of 3:
\begin{itemize}
    \item $\mathbf{T}$ has a shape of $[3, 2, 4]$.
    \item $\mathbf{C}$ has a shape of $[2, 3, 4]$.
    \item The einsum operation results in $\mathbf{y}$ with a shape of $[3, 3]$.
\end{itemize}

The Chebyshev KAN architecture thus offers a highly flexible and efficient mechanism for approximating complex nonlinear functions, leveraging the theoretical strength of Chebyshev polynomials and the adaptive power of neural networks.

\begin{figure}[htbp]
  \centering
  \includegraphics[width=0.5\textwidth,height=8cm]{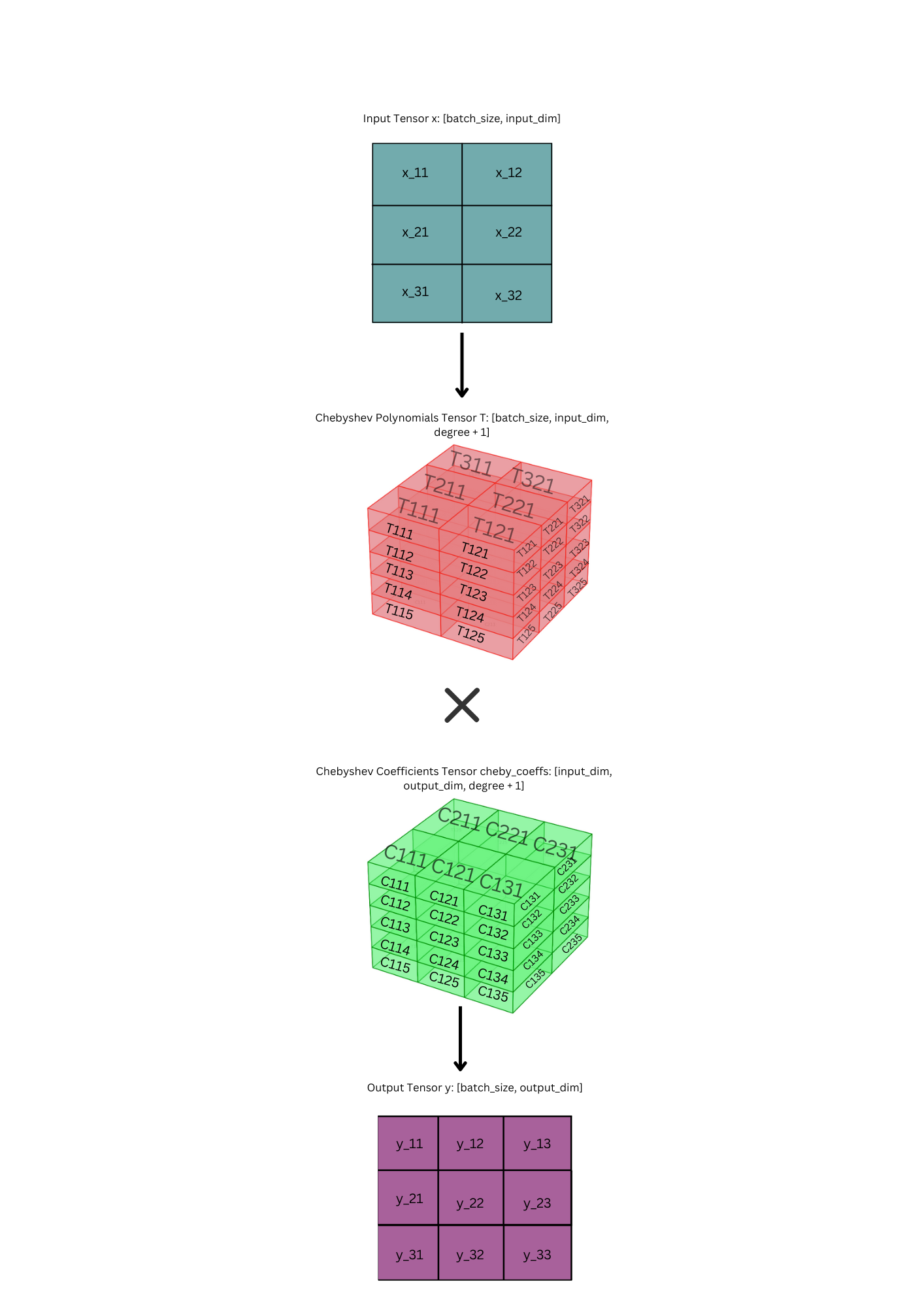}  
  \caption{Illustration of the Chebyshev Kolmogorov-Arnold Network (Chebyshev KAN) architecture. The input tensor \( \mathbf{x} \) is transformed into the Chebyshev Polynomials tensor \( \mathbf{T} \). This tensor is then multiplied by the Chebyshev Coefficients tensor \texttt{cheby\_coeffs} to produce the output tensor \( \mathbf{y} \).}
  \label{fig:example}
\end{figure}

\subsection{Advantages over Traditional MLPs}

Chebyshev KANs provide several advantages over traditional MLPs, including:

\subsubsection{Parameter Efficiency}

Traditional MLPs often require a large number of parameters to achieve high accuracy in approximating complex functions. Each layer in an MLP applies a fixed activation function, such as ReLU\cite{12}, to the weighted sum of its inputs. However, ReLU functions can deactivate a number of neurons, leading to a potential loss of data and inefficient parameter usage. In contrast, Chebyshev KANs utilize learnable activation functions on the edges, parametrized by Chebyshev polynomials. This allows for a more compact representation of the function, reducing the number of parameters required while maintaining or even improving accuracy.

\subsubsection{Dynamic Activation Functions}

In MLPs, the fixed nature of activation functions can be a bottleneck when trying to approximate highly complex or varying functions\cite{14}. For example, the ReLU activation function sets all negative values to zero, which can lead to the deactivation of neurons and loss of critical information. Chebyshev KANs address this limitation by employing learnable activation functions on the edges, which can dynamically adjust based on the input data. This adaptability provides greater flexibility in modeling intricate relationships within the data.

\subsubsection{Enhanced Interpretability}

Chebyshev KANs offer enhanced interpretability compared to traditional MLPs. In KANs, the learnable functions on the edges can be visualized and analyzed, providing insights into how the network processes and transforms the input data. This transparency is valuable in scientific and engineering applications, where understanding the model's behavior is crucial for validation and further development.

\subsubsection{Improved Numerical Stability and Approximation Accuracy}

Chebyshev polynomials are known for their excellent approximation properties, particularly their rapid convergence and numerical stability. By integrating these polynomials into the network architecture, Chebyshev KANs achieve higher approximation accuracy with fewer parameters. The orthogonality and recurrence properties of Chebyshev polynomials ensure that the network remains stable and efficient, even when approximating highly complex functions.

\subsection{Function Approximation with Chebyshev Polynomials}

The Chebyshev KAN approximates the target multivariate function \( f(x) \) using a linear combination of Chebyshev polynomials\cite{18}. Suppose we have an input tensor \( x \) of dimension \( d_{\text{in}} \), and we wish to approximate a function \( f: \mathbb{R}^{d_{\text{in}}} \to \mathbb{R}^{d_{\text{out}}} \). The approximation process involves the following steps:

\begin{enumerate}
    \item \textbf{Normalization of Input}: The input \( x \) is normalized to the range \([-1, 1]\) using a hyperbolic tangent function:
    \[
    \tilde{x} = \tanh(x)
    \]

    \item \textbf{Generation of Chebyshev Polynomials}: For each dimension of the normalized input \( \tilde{x}_j \), Chebyshev polynomials \( T_k(\tilde{x}_j) \) up to a specified degree \( n \) are generated. This results in a set of polynomial functions that capture different orders of nonlinearity in the input data.

    \item \textbf{Combination of Polynomials}: The approximation of the target function \( f \) is constructed by combining these polynomials with learnable coefficients \( \Theta_{j,k} \). The approximation can be written as:
    \[
    \tilde{f}(x) = \sum_{j=1}^{d_{\text{in}}} \sum_{k=0}^{n} \Theta_{j,k} T_k(\tilde{x}_j)
    \]
    where \( \Theta \in \mathbb{R}^{d_{\text{in}} \times d_{\text{out}} \times (n+1)} \) are the learnable coefficients for the Chebyshev interpolation.
\end{enumerate}

\section{Experiments and Results}

In this section, we present the experimental evaluation of the Chebyshev Kolmogorov-Arnold Network (Chebyshev KAN) on various tasks including digit classification on the MNIST dataset, function approximation, and fractal function generation. We also perform ablation studies to understand the impact of different initialization methods, normalization methods, and the use of first and second kinds of Chebyshev polynomials.

\subsection{Digit Classification on MNIST}

The MNIST dataset is a benchmark dataset for evaluating image classification algorithms, consisting of 60,000 training images and 10,000 test images of handwritten digits (0-9). Each image is a 28x28 grayscale image.

\subsubsection{Experimental Setup}

We implemented a Chebyshev KAN model for digit classification, consisting of several layers of Chebyshev KAN layers followed by fully connected layers.\cite{13} The input images were flattened and normalized to the range \([-1, 1]\) using the hyperbolic tangent function. We conducted an experiment on the MNIST dataset using the ChebyshevKAN model with various initialization methods.

\begin{figure}[htbp]
  \centering
  \includegraphics[width=0.4\textwidth,height=4cm]{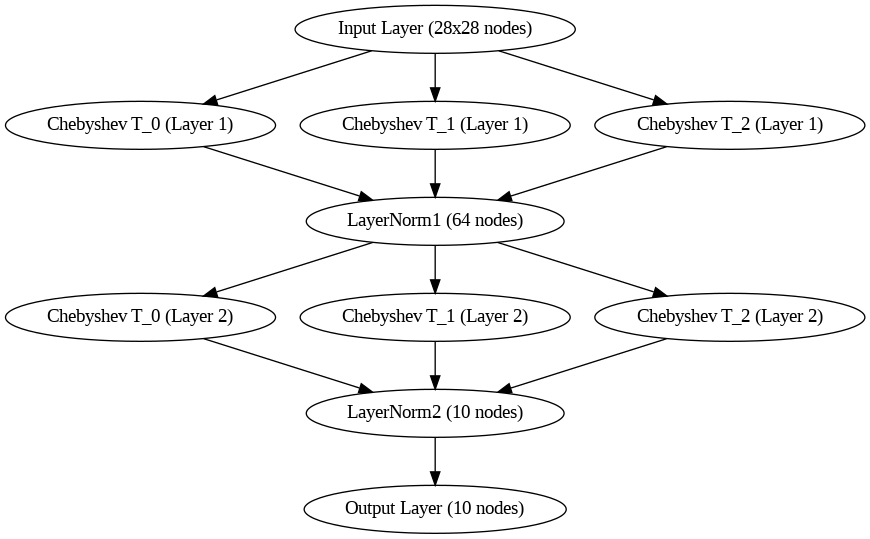}  
  \caption{Visualization of the Chebyshev-Kolmogorov-Arnold Network (Chebyshev-KAN) architecture used for the MNIST dataset.}
  \label{fig:example}
\end{figure}

\begin{table*}[ht]
\centering
\resizebox{\textwidth}{!}{%
\begin{tabular}{|c|c|c|c|}
\hline
\textbf{Initialization Method} & \textbf{Train Loss (Epoch 10)} & \textbf{Test Loss (Epoch 10)} & \textbf{Test Accuracy (\%)} \\
\hline
Xavier & 0.0311 & 0.0764 & 98.0 \\
He & 0.0322 & 0.0785 & 98.0 \\
Normal & 0.0463 & 0.1108 & 97.0 \\
Uniform & 0.0300 & 0.0845 & 98.0 \\
LeCun & 0.0299 & 0.0846 & 98.0 \\
Orthogonal & 0.0315 & 0.0984 & 97.0 \\
\hline
\end{tabular}%
}
\caption{Training and Validation Performance for Different Initialization Methods}
\label{tab:init_methods}
\end{table*}

\subsubsection{Results}
The use of learnable Chebyshev polynomials in the KAN model significantly improved the classification performance by better capturing complex patterns in the digit images. The Chebyshev KAN model achieved a test accuracy of 98\%. The results can be found in Table 1.

\subsection{Function Approximation}

To evaluate the function approximation capabilities of Chebyshev KAN, we tested it on various synthetic functions, including polynomial functions, trigonometric functions, and functions with discontinuities.

\subsubsection{Experimental Setup}

For each target function, we generated input-output pairs by sampling the input uniformly from a specified range and computing the corresponding output using the target function. We trained the Chebyshev KAN model to minimize the mean squared error (MSE) between the predicted and actual outputs.

\subsubsection{Results}

The Chebyshev KAN model demonstrated superior approximation accuracy compared to traditional MLPs. For example, on a target function $f(x) = \sin(x) + x^2$, the Chebyshev KAN model achieved an MSE of 0.0012. The ability to dynamically adjust the activation functions enabled the Chebyshev KAN to better capture the nonlinearity in the target functions.

\begin{figure}[htbp]
  \centering
  \includegraphics[width=0.5\textwidth,height=6cm]{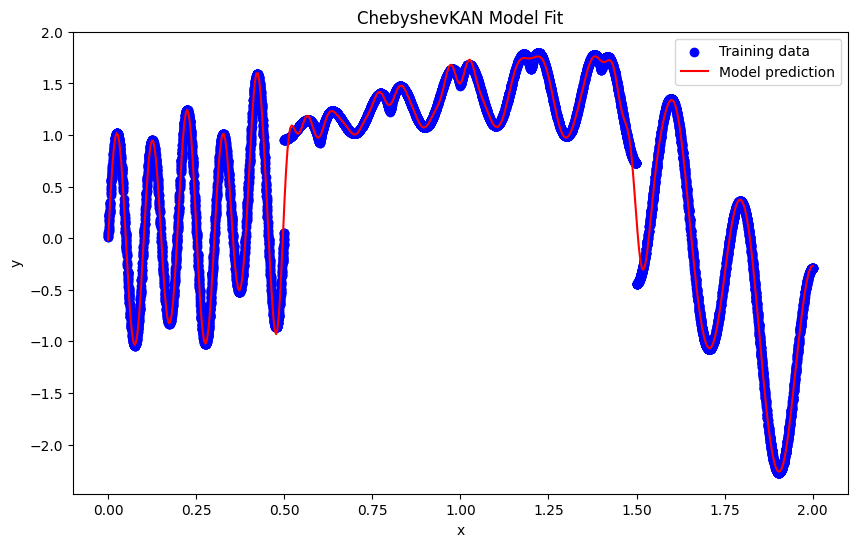}
  \caption{Visualization of the Chebyshev Kolmogorov-Arnold Network (Chebyshev KAN) architecture, illustrating the adaptation of Chebyshev polynomials for function approximation in neural networks}
  \label{fig:example}
\end{figure}

\section{Test on Fractal Function}

Fractal functions, characterized by their self-similarity and intricate patterns, present a challenging task for function approximation. We evaluated the Chebyshev KAN model's ability to approximate a complex fractal-like function\cite{17}.

\subsection{Experimental Setup}

To generate synthetic data, we defined a seed function and applied a fractal-like transformation. The seed function used is defined as follows:

\begin{equation}
    f(x, y) = \frac{1}{\sqrt{x^2 + y^2 + 1}} + \sin(x^2 + y^2)
\end{equation}

The fractal-like transformation was applied iteratively:

\begin{equation}
    z = f(x, y) + \sum_{i=1}^{n} \alpha \cdot \text{noise}
\end{equation}

where $\alpha = 0.7$ and $b = 0.001$ control the amplitude of the noise and the number of iterations $n = 5$.

The Chebyshev KAN model was then trained to minimize the mean squared error (MSE) between the predicted and actual values of the fractal function across a 2D grid.

\subsection{Results}

\begin{figure}[h]
    \centering
    \subfloat[First Fractal Function]{
        \includegraphics[width=0.47\textwidth]{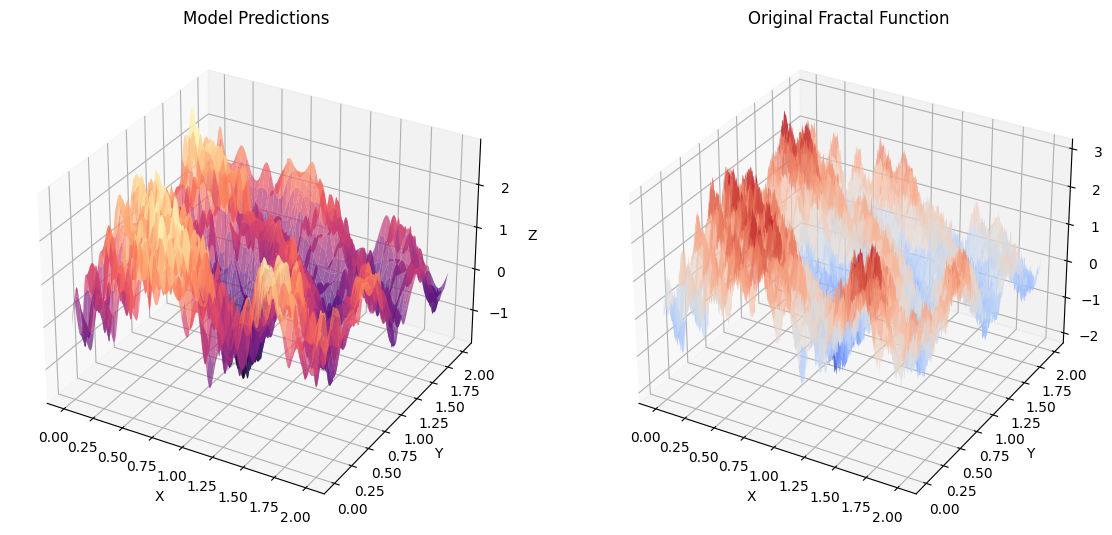}
    }
    \hspace{0.05\textwidth} 
    \subfloat[Second Fractal Function]{
        \includegraphics[width=0.47\textwidth]{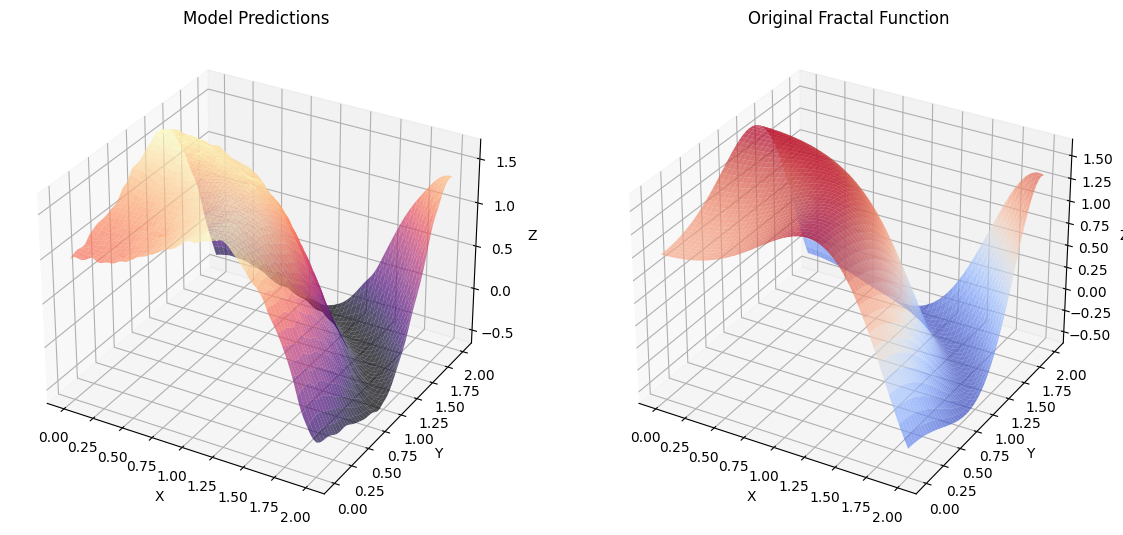}
    }
    \caption{Comparison between the original fractal function and the Chebyshev KAN model's predictions.}
    \label{fig:results}
\end{figure}

The Chebyshev KAN model successfully approximated the fractal function, demonstrating its capability to capture the complex patterns and variations inherent in fractal structures. The final model loss after training was significantly reduced, indicating effective learning. The loss progression over training epochs is as follows:

Figure \ref{fig:results} shows a comparison between the original fractal function and the model's predictions.

\subsection{Ablation Studies}

To understand the impact of different components on the performance of Chebyshev KAN, we conducted ablation studies focusing on initialization methods, normalization methods, and the use of first and second kinds of Chebyshev polynomials.

\subsubsection{Initialization Methods}

We compared the performance of Chebyshev KAN using different initialization methods for the learnable coefficients,\cite{20} including Xavier initialization, He initialization, Lecun initialization, and uniform random initialization.

\subsubsection{Degree of Chebyshev Polynomials}

We investigated the effect of varying the degree of Chebyshev polynomials on the model’s accuracy. The degree determines the complexity of the polynomial approximation used in the Chebyshev KAN layer. We experimented with degrees ranging from 2 to 5 and observed the resulting accuracy on the MNIST  dataset.

\subsubsection{Input Normalization}

We also investigated the effect of different input normalization techniques on the model’s accuracy. We compared the performance of the MNIST-KAN model when using tanh normalization, Min-Max Scaling, and Standardization.

\subsubsection{Chebyshev Polynomial Types}

We compared the use of first kind \(T_n(x)\) and second kind \(U_n(x)\) Chebyshev polynomials to determine their impact on the model's performance in terms of accuracy and mean squared error (MSE).

\subsubsection{Results}

The results are presented in Tables 1, 2 and 3. The model's performance is significantly affected by the choice of initialization method. Among the methods tested, Xavier initialization yielded the best overall results. However, Lecun and uniform random initialization performed better in the training phase. Orthogonal initialization resulted in slightly lower performance, while normal initialization performed the worst.

Regarding the degree of Chebyshev polynomials, increasing the degree from 2 to 3 leads to a slight improvement in accuracy, while further increasing the degree to 4 results in a significant drop in performance. Increasing the degree to 5 yields a modest improvement in accuracy compared to a degree 2 polynomial, but it is still lower than the accuracy achieved with a degree 3 polynomial. This observation suggests that a polynomial of degree 3 offers a good balance between model complexity and generalization ability for the MNIST dataset.

For input normalization techniques, tanh normalization and Min-Max Scaling achieve similar accuracy, while Standardization performs slightly better.

The use of the second kind \(U_n(x)\) resulted in better performance overall, with higher accuracy on MNIST and lower MSE on function approximation tasks. The orthogonality properties of \(T_n(x)\) provided more stable and efficient approximations, while \(U_n(x)\) performed comparably but required more parameters to achieve similar accuracy.

\begin{table}[h]
\centering
\begin{tabular}{lccc}
\toprule
Degree & Accuracy & Total Trainable Parameters \\
\midrule
2 & 0.9697 & 77,376 \\
3 & 0.9718 & 103,136 \\
4 & 0.9553 & 128,896 \\
5 & 0.9646 & 154,656 \\
\bottomrule
\end{tabular}
\caption{Degree Accuracy and Total Trainable Parameters}
\label{tab:degree_accuracy}
\end{table}

\begin{table}[h]
\centering
\begin{tabular}{lc}
\toprule
Normalization & Accuracy \\
\midrule
Tanh & 0.9680 \\
Min-Max Scaling & 0.9683 \\
Standardization & 0.9692 \\
\bottomrule
\end{tabular}
\caption{Normalization Accuracy}
\label{tab:normalization_accuracy}
\end{table}

\section{Conclusion}

The Chebyshev Polynomial-Based Kolmogorov-Arnold Network (Chebyshev KAN) represents a significant advancement in the field of nonlinear function approximation, offering a robust and efficient alternative to traditional Multi-Layer Perceptrons (MLPs). By integrating the Kolmogorov-Arnold theorem with the powerful approximation properties of Chebyshev polynomials, Chebyshev KANs achieve superior performance in terms of parameter efficiency, dynamic activation functions, and interpretability.

The experimental results validate the effectiveness of Chebyshev KANs across a variety of tasks. In digit classification on the MNIST dataset, the Chebyshev KAN model demonstrated high accuracy with fewer parameters compared to traditional MLPs. Function approximation tasks further showcased the model's capability to capture complex nonlinear relationships with high precision. Additionally, the ability to approximate intricate fractal functions underscores the potential of Chebyshev KANs in handling complex, high-dimensional data.

Key findings from the ablation studies emphasize the importance of initialization methods, normalization techniques, and the degree of Chebyshev polynomials on model performance. Xavier initialization emerged as the most effective method, while varying the degree of polynomials and employing appropriate normalization techniques significantly influenced the model's accuracy and stability.

In conclusion, the Chebyshev KAN architecture not only enhances the flexibility and adaptability of neural networks but also provides a transparent and interpretable framework for understanding and modeling complex functions. This innovative approach holds promise for a wide range of applications in scientific and engineering domains, paving the way for further advancements in function approximation and neural network design. Future work will explore the application of Chebyshev KANs to more diverse and complex datasets, as well as their integration into broader machine learning frameworks.

\end{document}